\documentclass[10pt,twocolumn,letterpaper]{article}

\usepackage{cvpr}              
\usepackage{xcolor}
\usepackage{times}
\usepackage{graphicx}
\usepackage{amsmath}
\usepackage{amssymb}
\usepackage{booktabs}
\usepackage{multirow}
\usepackage{makecell}
\usepackage{float}
\usepackage{bbding}
\usepackage{colortbl}
\usepackage[toc,page]{appendix}
\usepackage[accsupp]{axessibility}

\PassOptionsToPackage{prologue,dvipsnames}{xcolor}
\usepackage[dvipsnames]{xcolor}

\definecolor{cvprblue}{rgb}{0.21,0.49,0.74}
\usepackage[pagebackref,breaklinks,colorlinks,citecolor=cvprblue]{hyperref}

\definecolor{G}{HTML}{39b54a} 
\definecolor{R}{HTML}{ff0000} 
\definecolor{B}{HTML}{000000}
\newcommand\resdiff[3]{\textcolor{#1}{\scriptsize{~(${#2}${#3})}}}

\def\paperID{3159} 
\def\confName{CVPR}
\def\confYear{2024}

\title{Towards More Unified In-context Visual Understanding}

\author{Dianmo Sheng\textsuperscript{1}, 
\hspace{1em}Dongdong Chen\textsuperscript{2}, 
\hspace{1em}Zhentao Tan\textsuperscript{1}, 
\hspace{1em}Qiankun Liu\textsuperscript{3}, 
\hspace{1em}Qi Chu\textsuperscript{1}, \\
 \hspace{1em}Jianmin Bao\textsuperscript{2}, 
 \hspace{1em}Tao Gong\textsuperscript{1}\thanks{Tao Gong is the corresponding author}, 
 \hspace{1em}Bin Liu\textsuperscript{1}, 
 \hspace{1em}Shengwei Xu\textsuperscript{4}, 
 \hspace{1em}Nenghai Yu\textsuperscript{1} \\
\textsuperscript{1}School of Cyber Science and Technology, University of Science and Technology of China \\
Anhui Province Key Laboratory of Digital Security \\
the CAS Key Laboratory of Electromagnetic Space Information \qquad \textsuperscript{2}Microsoft Research \\
\qquad \textsuperscript{3}Beijing Institute of Technology\qquad \textsuperscript{4}Beijing Electronic Science and Technology Institute \\
}
\begin{document}


\maketitle
\begin{abstract}
\vspace{-0.6em}
The rapid advancement of large language models (LLMs) has accelerated the emergence of in-context learning (ICL) as a cutting-edge approach in the natural language processing domain. Recently, ICL has been employed in visual understanding tasks, such as semantic segmentation and image captioning, yielding promising results. However, existing visual ICL framework can not enable producing content across multiple modalities, which limits their potential usage scenarios. To address this issue, we present a new ICL framework for visual understanding with multi-modal output enabled. First, we quantize and embed both text and visual prompt into a unified representational space, structured as interleaved in-context sequences. Then a decoder-only sparse transformer architecture is employed to perform generative modeling on them, facilitating in-context learning. Thanks to this design, the model is capable of handling in-context vision understanding tasks with multimodal output in a unified pipeline.Experimental results demonstrate that our model achieves competitive performance compared with specialized models and previous ICL baselines. Overall, our research takes a further step toward unified multimodal in-context learning. 
\end{abstract}  
\vspace{-0.5cm}
\section{Introduction}
\label{sec:intro}
With the rapid progress of large language models, \textit{in-context learning (ICL)}~\cite{GPT-3,Metaicl,selfadaICL} has gradually become a new paradigm in the field of natural language processing (NLP). As introduced in GPT-3~\cite{GPT-3}, given language sequences as a universal interface, the model can quickly adapt to different language-centric tasks by utilizing a limited number of prompts and examples. 

\begin{figure*}[t]
   \centering
   \includegraphics[width=0.99\linewidth]{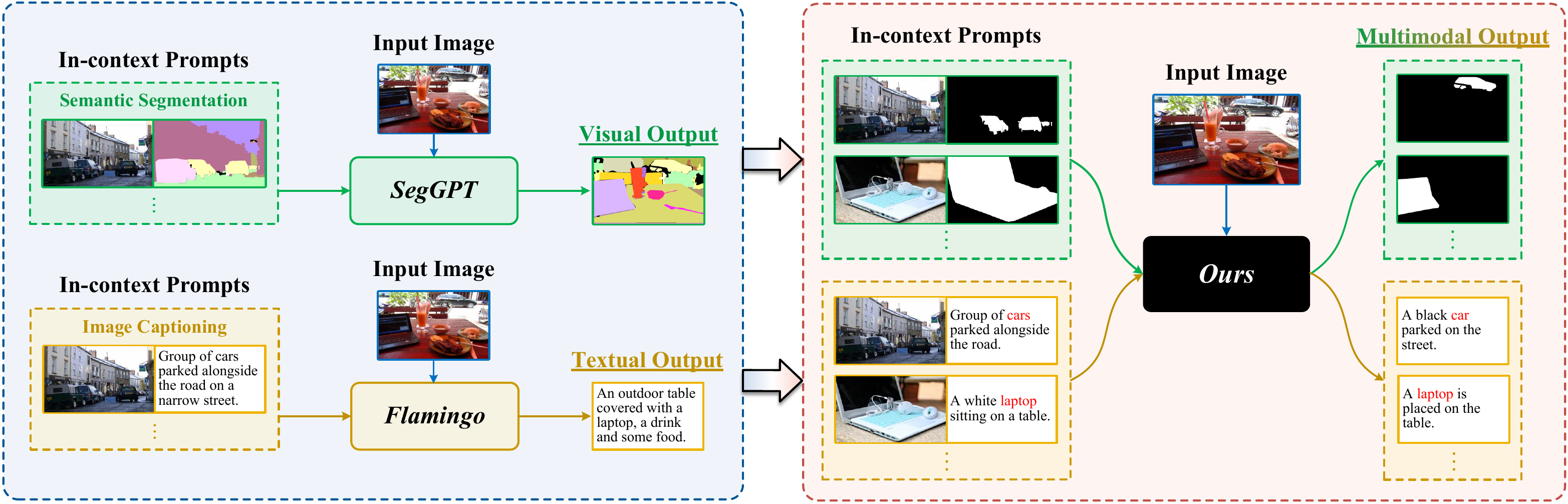}
     \vspace{-1mm}
   \caption{Motivation illustration of our method. In earlier efforts, existing in-context visual understanding models were confined to a particular output modality. For instance, SegGPT specialized in ``\textbf{\textcolor[RGB]{0,153,69}{Image} $\rightarrow$ \textcolor[RGB]{0,153,69}{Image}}'' applications, tailored for tasks involving image segmentation. Similarly, Flamingo was purpose-built for ``\textbf{\textcolor[RGB]{0,153,69}{Image} $\rightarrow$ \textcolor[RGB]{204,153,0}{Text}}'' scenarios, focusing on language-centric tasks such as image captioning. In contrast, we take a further attempt to design a unified model capable of handling multimodal in-context visual understanding tasks for ``\textbf{\textcolor[RGB]{0,153,69}{Image} $\rightarrow$ \textcolor[RGB]{0,153,69}{Image} / \textcolor[RGB]{204,153,0}{Text}}'' scenarios.}
  \label{fig:motivation}
 \end{figure*}

Some following works~\cite{Flamingo,frozen} present some early attempt at applying ICL into the vision-language (VL) tasks with the design of interleaved image and text data. For example, Flamingo~\cite{Flamingo} takes the image input as a special ``$\textless$image$\textgreater$'' token to conduct the interleaved input prompt as text, and injects visual information into pre-trained LLMs with gated cross-attention dense block. It demonstrates a remarkable capability to address various vision-language tasks. However, the language-only LLM decoder design makes it only able to output text outputs. 

More recently, some works start to apply the similar ICL idea into the vision-only tasks via formulating the learning goal as image inpainting~\cite{VP,Painter,Seggpt}. With the well-collected multi-task vision datasets and unified grid image prompt design, these works utilize pre-trained masked image modeling models to give a perspective of what can be general-purpose task prompts in vision. For instance, SegGPT~\cite{Seggpt} studies the fundamental visual understanding problem, segmentation task, as an in-context coloring problem to achieve the in-context segmentation capability. Yet, the pre-trained vision-centric inpainting framework confines the output modality to be image only.
Therefore, a straightforward question is ``\textit{How to perform in-context learning with multimodal output enabled for visual understanding in a unified framework}?"

Standing on the shoulders of predecessors, in this paper, we present the first attempt at multimodal in-context learning. The central concept aims to unify vision-language data via modality-specific quantization and shared embedding, then perform next-token prediction on the well-organized interleaved sequences of in-context samples.

In detail, we first develop detailed and comprehensive vision and language prompts, carefully designed to represent various vision understanding tasks. Then we employ modality-specific quantizers to transform the formatted in-context prompts and the visual input into discrete tokens respectively. Following this, a unified embedding layer is used to map these tokens into a shared representational space. Once the model outputs prediction tokens with specific prompts, the modality-specific decoders automatically decode them into the intended domains. This design effectively allows for multimodal input and output. To facilitate the in-context learning on unified representations, we further combine the autoregressive transformer with the Mixture of Experts (MoEs). The autoregressive transformer produces a natural contextual association based on the next-token prediction, while MoEs \cite{Gshard,Switch-transformer} serve as a promising solution for multi-task learning by dynamically activating sub-networks without the need for task-specific modules. 
Following previous in-context prompts formats, we take semantic segmentation and dense captioning as the example image understanding tasks, and formatting semantic category information as the clue across multiple in-context samples. Through extensive experiments and analysis, we demonstrate that our model can facilitate in-context learning on vision understanding tasks and enable multimodal outputs within a unified model. 

\section{Related Works}
\label{sec:relatedwork}
\noindent\textbf{In-Context Learning.}
As the dimensions of both model size and corpus size escalate~\cite{Bert,GPT-2,GPT-3,Palm}, large language models (LLMs) exhibit an aptitude for in-context learning (ICL), namely, the capacity to distill knowledge from a limited array of contextual examples. GPT-3~\cite{GPT-3}, for instance, pioneers the articulation of various natural language processing (NLP) tasks as text completion conundrums, a strategy predicated on the provision of prompts and examples. This novel methodology considerably simplifies the integration of task knowledge into LLMs by modifying the demonstrations and templates, a concept substantiated by various studies~\cite{fantastically,selfadaICL,CoT}. 

Within the field of computer vision, the study~\cite{VP} initially advances an in-context training paradigm utilizing image inpainting on illustrations and infographics derived from vision-related literature, which shows competencies in fundamental CV tasks. Additionally, the study by Painter~\cite{Painter} employs masked image modeling on continuous pixels to conduct in-context training with self-organized supervised datasets in seven tasks, and yields highly competitive outcomes on them. Subsequently, SegGPT~\cite{Seggpt} is a dedicated method trying to solve diverse and unlimited segmentation tasks with a similar framework. Recent studies have concentrated on how to enhance the ICL capability in vision, such as prompt selection~\cite{prompt-SelF} and the execution of nearest neighbor retrieval utilizing a memory bank~\cite{Hummingbird}.

Prior works have typically been confined to specific domains. In contrast, our study is conducted across both vision and language domains, as we aspire to realize the potential of multimodal in-context learning.

\vspace{0.5em}
\noindent\textbf{Multimodal Understanding and Generation.}
Multimodal understanding and generation represent an emerging frontier in artificial intelligence that seeks to interpret and synthesize information across various forms of data, such as text, images, sounds, and even more modalities. Inspired by the success of ChatGPT as well as GPT-4~\cite{chatgpt,gpt4}, recent works primarily concentrate on aligning visual features with the pre-trained LLMS for multimodal comprehension tasks~\cite{minigpt4,BEiT-3,llava,mplug-owl,kosmos1,blip2,visionllm,otter}. While pre-trained LLMs have empowered systems to follow human instructions for vision-language interactions, their application has been confined to generating textual outputs.

Expanding the horizons of multimodal capabilities, a burgeoning spectrum of studies~\cite{imagebind,codi,gill,next-gpt,emu,cm3leon} are pioneering innovations in both understanding and generative capacities across modalities. IMAGEBIND~\cite{imagebind} utilizes the image-paired data to connect five different modalities with a single joint embedding space, demonstrating impressive zero-shot capabilities across these modalities. Otherwise, CoDi~\cite{codi} introduces a composable generation strategy by bridging alignment in the diffusion process, facilitating the synchronized generation of any combination of output modalities, including language, image, video, or audio. Furthermore, NExT-GPT~\cite{next-gpt} integrates an LLM with multimodal adaptors and diverse diffusion decoders, enabling it to perceive inputs and generate outputs in arbitrary combinations of text, images, videos, and audio with understanding and reasoning.

However, these models are not designed for in-context learning, without the benefit of the multiple prompts.

\vspace{0.5em}
\noindent \textbf{Mixture of Experts models.}
Mixture of Experts (MoEs), which have demonstrated remarkable success in both computer vision \cite{wang2020deep,V-MoE,lou2021cross} and natural language processing \cite{shazeer2017outrageously,kudugunta2021beyond,Switch-transformer,Glam,St-moe} with the context of conditional computation. Conditional computation aims to increase the number of model parameters without significantly increasing computational cost by selectively activating relevant parts of the model based on input-dependent factors \cite{chen1999improved,davis2013low}. \cite{shazeer2017outrageously} first provides compelling evidence for the efficacy of MoEs by incorporating MoE layers into LSTM models. Building upon this, subsequent studies \cite{Mesh,Gshard,Switch-transformer,kim2021scalable} extend the application of this approach to transformer architectures. 

With different routing strategies, MoE models have also been studied for multitask learning \cite{kudugunta2021beyond,Dselect-k,Uni-perceiver-moe} and multimodal learning \cite{LIMoE,VL-MoE} as well. Recent work VL-MoE~\cite{VL-MoE} is the first work to combine modality-specific MoEs with generative modeling for vision-language pretraining. In this work, we further study the potential of combining autoregressive transformer with MoE for vision-language in-context learning. 

\section{Method}
\label{sec:method}
In this section, We present a multimodal in-context framework that can seamlessly integrate the strengths of language models with the specific requirements of vision-language tasks for in-context learning. We first introduce well-organized vision-language prompts to describe foundational visual understanding tasks like segmentation and captioning (Section~\ref{sec:vl_prompt}). After conducting the input into predefined prompts format, we quantize in-context prompts with the input pair into discrete codes using modality-specific tokenizers, and then embed them into unified representations with a general embedding network (Section~\ref{sec:uni-repre}). Then a decoder-only transformer with sparse MoEs is introduced to perform generative modeling on the interleaved unified representations (Section~\ref{sec:model}). In the following paragraph, we will elaborate on each part in detail.

\begin{figure*}[t]
   \centering
   \includegraphics[width=0.86\linewidth]{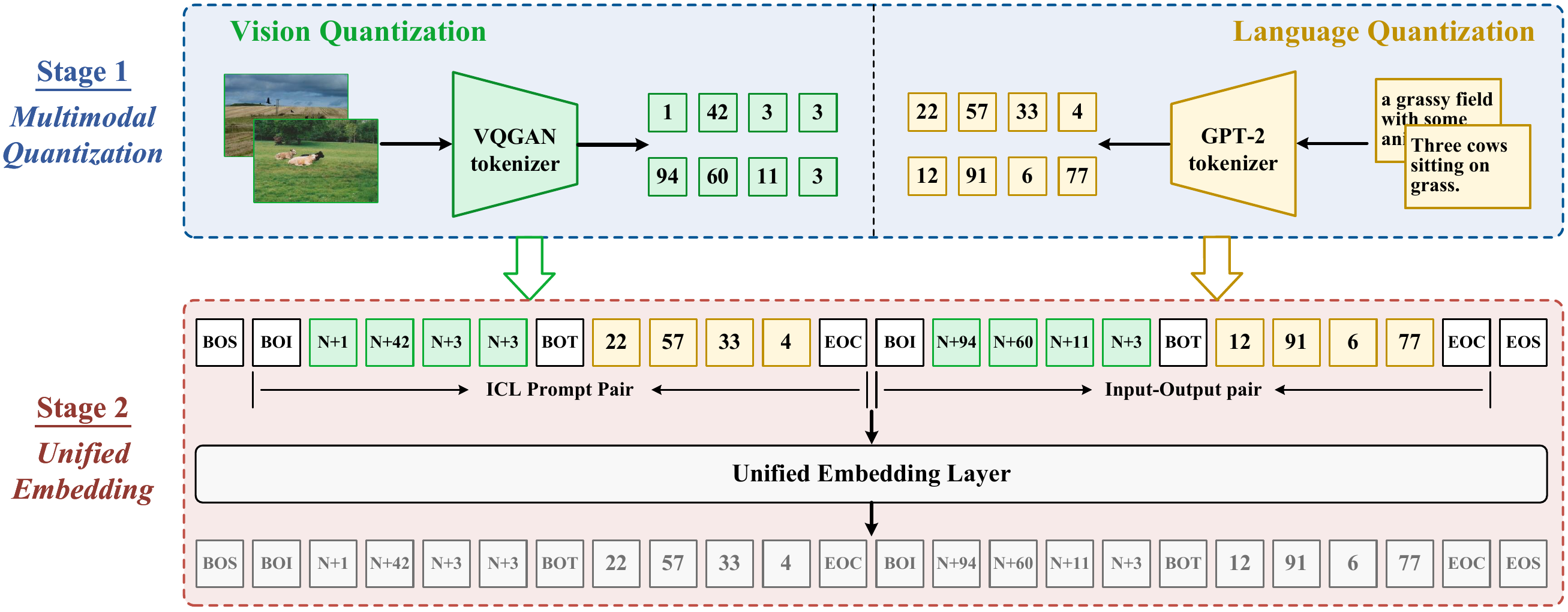}
   \caption{Overview of our unified multimodal representations pipeline with two stages. During the multimodal quantization phase, visual and linguistic inputs are encoded into discrete tokens via modality-specialized tokenizers: specifically, VQGAN's tokenizer for visual data and GPT-2's tokenizer for texts. After that, in the unified embedding stage, multimodal discrete tokens are formatted as an interleaved sequence with special tokens. Then a unified embedding layer projects the sequence into general representations.}
   \label{fig:uni_repre}
 \end{figure*}

\subsection{Vision-Language Prompt Design}
\label{sec:vl_prompt}
We begin by implementing unified vision-language prompts to depict different types of vision-language tasks. We treat $k$ in-context samples with input and output like "$(i_1, o_1), \cdots , (i_{k+1}, o_{k+1})$" as interleaved data, and embed them in the discrete token space. This innovative design provides the flexibility required for customizing vision or vision-language tasks according to specific needs and preferences.


\vspace{0.5em}
\noindent \textbf{Vision-Only Tasks.}
Following previous works, we conduct all vision-only tasks as an inpainting task. However, the inpainting is performed in token space. For every image pair that is composed of an original image and its corresponding task output, we first quantize them into discrete tokens utilizing a pre-trained image quantizer. A special tag ``[BOI]" is inserted in front of each image's token representation. Then we concatenate each pair's visual tokens obeying the order of precedence. This structure creates a cohesive relationship between the two in-context pairs, framing them both as visual token components.

\vspace{0.5em}
\noindent \textbf{Vision-Language Tasks.}
For vision-language tasks, here we take the dense captioning task as an example. The prompts are clear and closely resemble those of natural language processing (NLP) tasks. Similar to existing methods~\cite{Flamingo}, multiple captioning samples can be treated as interleaved image and text data. For each image, we quantize them the same way as in vision-only tasks, with the special ``[BOI]" tag. For the text part, we describe the region caption with corresponding instance category and bounding box (bbox) like ``\textit{Category: $\textless c\textgreater$. Bboxes: [$x_1, y_1, x_2, y_2$]. Caption: $\textless text\textgreater$.}" While $P=\{x_i, y_i\}^N_{i=1}$ represents points that locate the object. $\textless text\textgreater$ represents the placeholder of caption tokens. We also add a special tag ``[BOT]" at the beginning of each caption. After being tokenized by looking up the vocabulary, we use a similar concatenation strategy to get the in-context token representations. 

At the conclusion of each segment of in-context tokens, we incorporate an ``[EOC]" tag to signify the completion of in-context samples.
\vspace{-1mm}
\subsection{Unified Multimodal Representations.}
\label{sec:uni-repre}
Building upon the foundation of multimodal in-context prompts discussed in Section~\ref{sec:vl_prompt}, how to facilitate the model understanding multimodal input in a unified manner is a challenging problem. Revisiting previous vision-language models~\cite{frozen, Flamingo}, we decide to utilize the discrete token method as the bridge between the various input and the model embedding space. In this section, we will demonstrate the preparation for a general training recipe with multimodal in-context inputs by unifying representations based on modality-specific quantization. 

\vspace{0.5em}
\noindent \textbf{Multimodal Quantization Stage.}
We leverage existing well-known modality-specific quantizers to encode multimodal data into discrete tokens. 
As illustrated in Figure~\ref{fig:uni_repre}, for image data, we adopt the vector quantizer used in VQGAN~\cite{VQGAN}. Given an image $x_{img} \in \mathbb{R}^{H\times W\times 3}$, the quantization step is performed by searching the nearest embedding in the learned, discrete codebook $\mathcal{Z} = \{z_k\}^K_{k=1} \subset \mathbb{R}^{n_z}$, where $n_z$ is the codebook size, which can be formulated as:
\begin{equation}
    z_{q\_i} = \mathop{\arg\min}\limits_{z_k\in\mathcal Z}\|E(x_{img})-z_k\|_2.
\end{equation}
where $z_{q\_i}$ is the quantized encoding of $x_{img}$, and $E$ represents for the convolution encoder. We add the visual tokens to the text vocabulary. 

For the text part, the subword Byte-Pair Encoding (BPE) tokenizer in GPT-2~\cite{GPT-2} is utilized.
In the context of encoding information, BPE tokenizer quantizes $x_{text}$ into tokens $z_{q\_t}$ by looking up the vocabulary. We treat the category label $c$ as the natural language format, with two special tags $\rm \textless c\_st\textgreater$ and $\rm \textless c\_ed\textgreater$ denoting the start and end of this part. Compared with the class tokens proposed in~\cite{visionllm}, category label in language offers the potential for generalization to unseen classes. For the bbox information, we adopt a similar method in~\cite{Pix2Seq2}. After normalizing the coordinates $P$ with 3 decimal places according to the size of the image, we map it to predefined tokens $\rm \{ \textless bin\_0\textgreater, \cdots, \textless bin\_1000\textgreater \}$. Additional start and end tags $\rm \textless b\_st\textgreater, \textless b\_ed\textgreater$ are placed at both ends of the bbox. Therefore, we can control the precision of coordinates with fewer tokens than the numerical representation. 

\vspace{0.5em}
\noindent \textbf{Unified Embedding Stage.}
After quantizing each modality data into discrete tokens, we take the embedding step. Here, we treat data in both modalities equally, as all the tokens will be mapped into a unified representation embedding space by a linear layer. Then, all in-context token embeddings will be concatenated sequentially as ``$(z^1_{q\_i},z^1_{q\_t}),\cdots,(z^{k+1}_{q\_i},z^{k+1}_{q\_t})$'' and fed into the model. This design offers generality and scalability for multimodal knowledge transfer. Thus, the model can handle interleaved image and text inputs like Flamingo~\cite{Flamingo}.

\subsection{Model Architecture and Training Objective}
\label{sec:model}
\begin{figure}[t]
   \centering
   \includegraphics[width=0.99\linewidth]{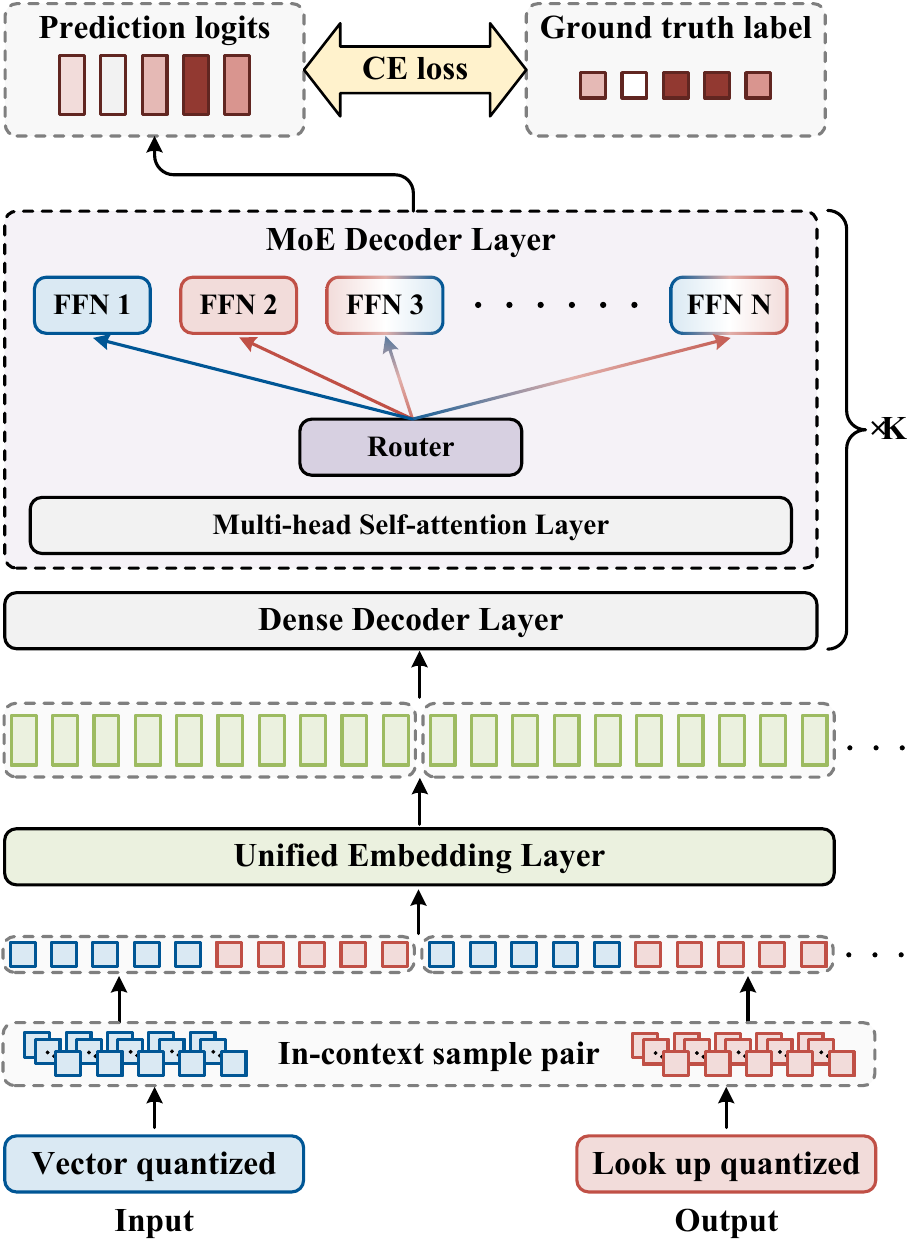}
   \caption{Overview of our pipeline. Here, we take the CA-ICL captioning task as an example. Multiple in-context samples and the input pair are first tokenized using modality-specific tokenizers and then projected into unified embedding representations. After undergoing interleaved concatenation, the tokens are inputted into the model for generative modeling.}
   \label{fig:model}
 \end{figure}

After the unification of various modality data, we are now going to discuss how to perform in-context learning in a general framework.
We construct our model using a GPT-2 style decoder-only transformer architecture with the sparse MoEs for multimodal in-context learning. As shown in Figure~\ref{fig:model}, the overall framework is very simple and straightforward. With the interleaved input representations, we utilize next-token prediction for modeling the contextual information. The model's predictive logits will undergo a sampling process to convert them back into tokens, which are subsequently decoded by the respective tokenizer of each modality. Consequently, the model can achieve multimodal input prompts and prediction, rather than being limited to specific output domains owing to the pre-trained backbone. 

\vspace{0.5em}
\noindent \textbf{Attribute Routing MoE.}
Different tasks with shared parameters may conflict with each other as described in previous works~\cite{Uni-perceiver-moe,Switch-transformer}. To mitigate the task interference issue, we utilize MoE layers, which allow different modalities and tasks to use separate parameters. For details, we replace the FFN block in each MoE decoder layer with the sparse MoE layer with $N$ experts introduced in~\cite{V-MoE}. Following Uni-Perceiver-MoE, we adapt the attribute routing strategy for in-context tokens, and top-k gating is implemented to decide the gating decision for the embedding of each token $x \in \mathbb{R}^D$. Therefore the calculation of gating is formulated as: $\mathcal{G}(x) = \mathrm{top}_k(\mathrm{softmax}(W_g(x)))$, where $W_g$ is the learnable weights of the router, and $\mathrm{top}_k(\cdot)$ represents operator that choose the largest $k$ values. After gating, the output of sparse MoE layer is the weighted combination of the activated experts' computation: $x_{out} = \sum_{i=1}^{N}\mathcal{G}(x)_i \cdot \mathrm{FFN}_i(x)$. 

\vspace{0.5em}
\noindent \textbf{Loss Function.}
Unlike previous vision generalists~\cite{VP,Painter,Seggpt} using masked image modeling as the learning objective, we perform generative modeling on interleaved in-context representations like Flamingo~\cite{Flamingo}, benefiting from the natural context understanding by leveraging next token prediction. 

The cross-entropy loss is employed on the output tokens of each in-context pair as well as the input pair, which constrains the similarity between model predictions $\mathcal{P}_{pred}$ and ground-truth tokens $\mathcal{P}_{gt}$, represented as: 
\begin{equation}
\mathcal{L}_{out} = \sum_{i=1}^{k+1}{\rm CE}(\mathcal{P}_{pred}^i, \mathcal{P}_{gt}^i)
\end{equation}

We also utilize the auxiliary loss introduced in GShard~\cite{Gshard} to optimize the gating network of MoEs, and the whole loss function can be represented as:
\begin{equation}
\mathcal{L} = \mathcal{L}_{out} + \lambda \cdot \mathcal{L}_{aux}
\end{equation}
where $\lambda$ is the weight of auxiliary loss.

\section{Experiments}
\label{sec:exp}
\subsection{Datasets and Benchmarks.}
\label{sec:data}
Prior works in visual in-context learning predominantly aimed to integrate concepts from NLP into conventional visual tasks. As detailed in MAE-VQGAN~\cite{VP}, Painter~\cite{Painter} and SegGPT~\cite{Seggpt}, each task involves creating a grid-structured image. However, these approaches overlook task-specific comprehension, merging all tasks into a singular prompt. Consequently, we propose a redefined approach to traditional visual tasks with semantic clues, emphasizing vision-language understanding tasks such as semantic segmentation and image captioning, which are named class-aware in-context (short for CA-ICL) segmentation and captioning respectively. 

\begin{figure}[t]
   \centering
   \includegraphics[width=\linewidth]{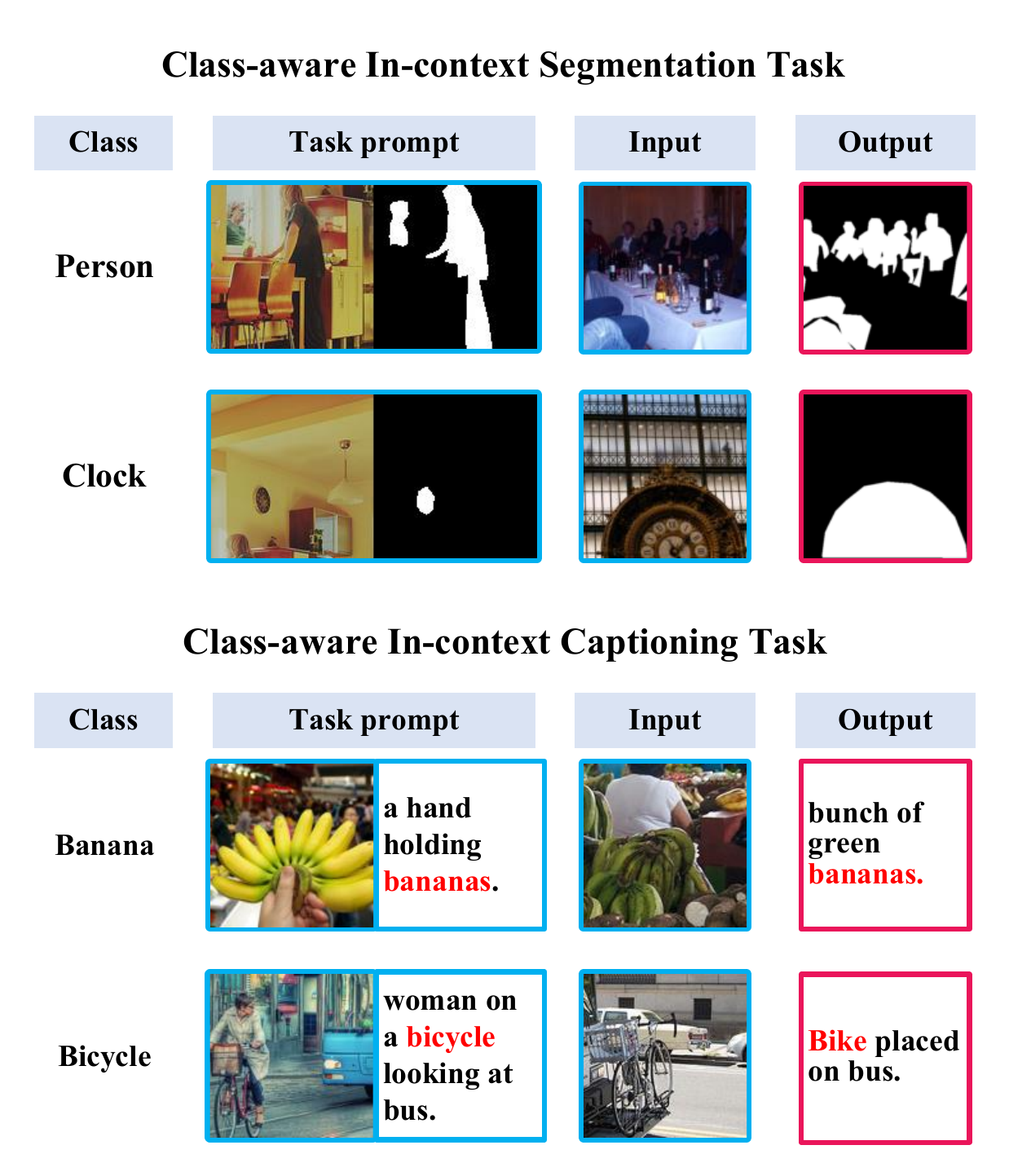}
     \vspace{-2mm}
   \caption{Class-aware in-context understanding task definitions. For the sake of easy demonstration, only one in-context sample is used here. The blue boxes \textcolor{cyan}{$\square$} on the left display the inputs of the model, while the red boxes \textcolor{magenta}{$\square$} on the right show the corresponding output. (In the absence of additional clarification, subsequent notations convey the same meaning.)}
   \label{fig:ca_icl}
 \end{figure}

\vspace{0.5em}
\noindent \textbf{CA-ICL Segmentaion.} 
As depicted in Figure~\ref{fig:ca_icl}, for segmenting instances of a particular class, each in-context sample is provided solely with the desired class segmentation mask. We conduct the data with the entire MS-COCO dataset, which contains 80 object classes. For each category, a mask pool is built for in-context sampling. Finally, we collect about 350k class masks for training and 15k class masks for validation. \textbf{Evaluation Metric:} We take the conventional semantic segmentation metric Mean Intersection over Union (MIoU) for evaluation. Given that the output is a binary mask, we also present the Mean Absolute Error (MAE) scores.

\vspace{0.5em}
\noindent \textbf{CA-ICL Captioning.} 
For the CA-ICL captioning, we also take the class information as the in-context clue, with each in-context sample containing the caption for the desired category. Here, we use the Visual Genome dataset, from which each image has multiple annotations, including object labels and caption annotations for each region of the image. We selectively use categories that correspond with those in the MS-COCO dataset, ensuring that each class has more than 100 descriptions. Finally, we collected about 460k region descriptions for training and 2k region descriptions for the test set. \textbf{Evaluation Metric:} Captioning performance is assessed using the BLEU4, METEOR, and CIDEr metrics, which are standard in image captioning tasks. When incorporating bbox information in prompts, we also present the mean Average Precision (mAP) metric following~\cite{densecap}. By filtering the prediction with predefined thresholds on IoU and METEOR, the average of the APs obtained for all pair-wise combinations of the two thresholds to evaluate both localization and description accuracy.

\vspace{0.5em}
\subsection{Implementation Details.}
\label{sec:implementation}
For the image tokenizer, we adopt VQ-GAN tokenizer~\cite{VQGAN} with a vocabulary size of 1024 and 16x downsampling ratio, which is pre-trained on the Imagenet dataset. The input image resolution is set to $256\times256$, leading to 256 tokens after quantization. For the text tokenizer, we employ GPT-2 BPE tokenizer~\cite{GPT-2} with a vocabulary size of 50257. We implement our model with GPT-small model architecture while replacing the FFN in part of the decoder layers with attribute routing MoEs introduced in~\cite{Uni-perceiver-moe}. 
Please refer to the supplementary for detailed architecture hyperparameters. 

During each training iteration, the number of in-context samples is set to 3 by default. All parameters are trained from scratch. The weight $\lambda$ is set to 0.02. For optimization, we employ the AdamW algorithm with a base learning rate of 1e-4, complemented by a weight decay of 0.05. We utilize gradient clipping at a value of 0.5 to stabilize the training process, ensuring consistent performance throughout. Unless otherwise specified, the training runs for 40 epochs with a batch size of 512 on 8 NVIDIA A6000 GPUs.

\subsection{Ablation Studies}
\label{sec:ablation}
In this section, we conduct an ablation study of our method from three perspectives: task definition, model definition, and multi-task co-training strategy. Without additional statements, the experiments are conducted using images in 128 resolution with 20 epochs of training.

\begin{table}[t]
  \small
  \centering
  \begin{tabular}{cc|cc}
  \toprule
   diverse sizes & large scale & MIoU \textcolor{red}{$\uparrow$}  & MAE \textcolor{red}{$\downarrow$} \\
    \midrule
    \XSolidBrush        & \XSolidBrush        & 31.82        & 0.176  \\
    \Checkmark          & \XSolidBrush        & 33.54        & 0.172  \\
    \XSolidBrush        & \Checkmark          & 42.87        & 0.133  \\
    \rowcolor{blue!10} \Checkmark          & \Checkmark          & \textbf{45.04}        & \textbf{0.128}  \\
  \bottomrule
\end{tabular}
\caption{Ablation of object size and scale in class-aware in-context segmentation task. Regarding object size, we adopt the MS-COCO definition, for whether to include small instances with an object area less than $32^2$ square units. For object scale considerations, the crop region is taken into account. The highlighted row indicates the best choice. (In the absence of additional clarification, subsequent notations convey the same meaning.)}
\label{tab:ca_icl_segmentation}
\end{table}

\begin{table}[t]
  \small
  \centering
  \begin{tabular}{cc|cc}
  \toprule
   bbox\_image & bbox\_text &  B@4 \textcolor{red}{$\uparrow$}  & CIDEr \textcolor{red}{$\uparrow$}\\
    \midrule
    \XSolidBrush        & \XSolidBrush        & \textbf{7.9}        & 104.4  \\
    \Checkmark          & \XSolidBrush        & 0.0        & 2.7  \\
    \rowcolor{blue!10} \XSolidBrush        & \Checkmark          & 7.8        & \textbf{112.0}  \\
  \bottomrule
\end{tabular}
\caption{Ablation study on the impact of bbox information in class-aware in-context caption task. ``bbox\_image'' and ``bbox\_text'' indicate that the bounding box is in image type or in text format.}
\label{tab:ca_icl_caption}
\end{table}

\begin{figure}[t]
   \centering
   \includegraphics[width=\linewidth]{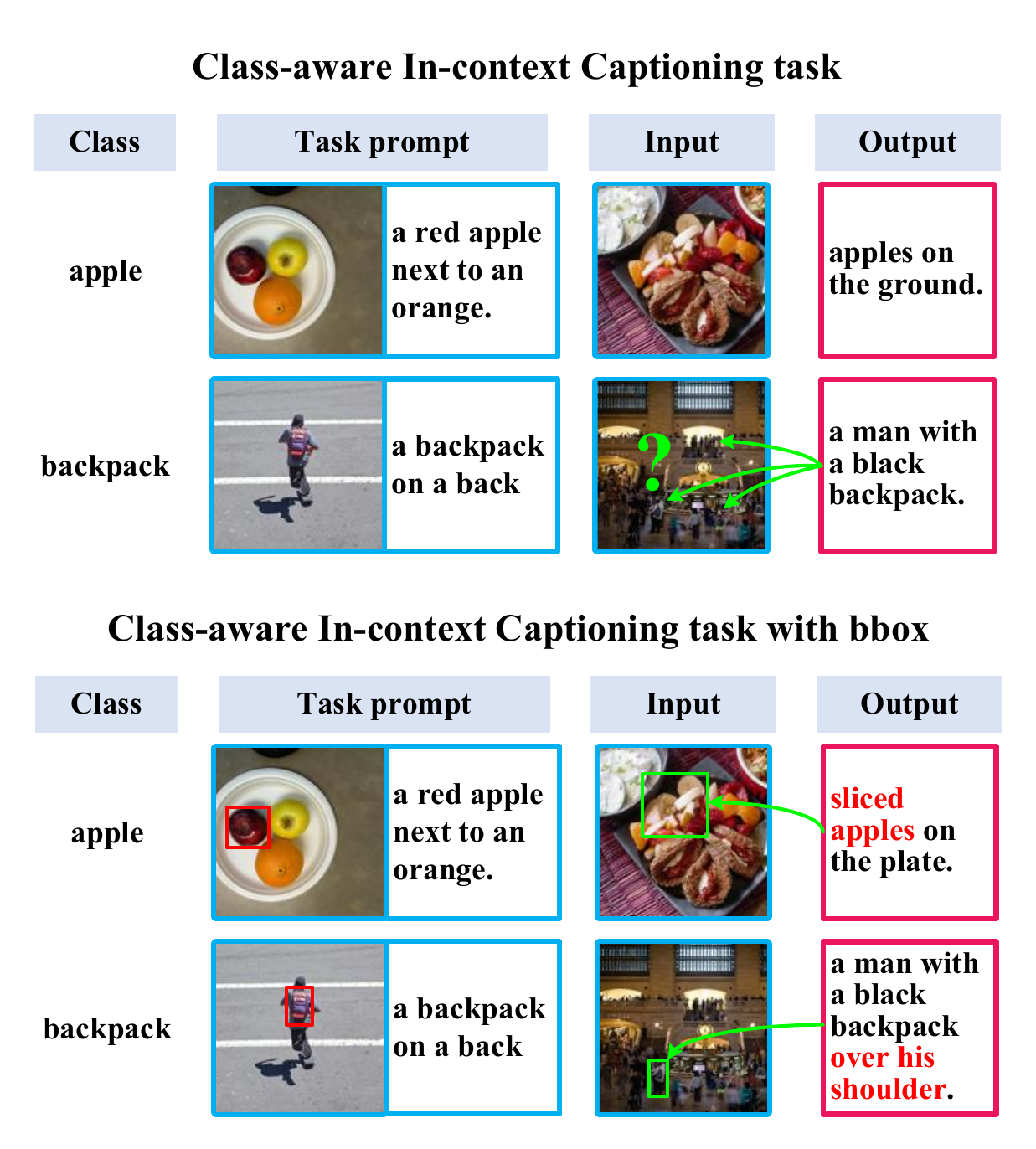}
   \vspace{-6mm}
   \caption{Analysis of the impact of including bbox information. For better visualization, the ground truth bboxes are indicated by rose boxes \textcolor{magenta}{$\square$}, while the predicted bboxes are highlighted in green boxes \textcolor{green}{$\square$}. With the bbox information in prompts, the model yields more precise descriptions that are aligned with the specified region locations.}
   \label{fig:ca_icl_cap_bbox}
 \end{figure}

\vspace{0.5em}
\noindent \textbf{Class-aware In-context Task Definitions.}
\label{sec:task_definition}
In our exploration of two proposed in-context learning tasks, we rigorously examine the task definitions. As demonstrated in Table~\ref{tab:ca_icl_segmentation}, we investigate the object size and scale within each in-context sample for the CA-ICL segmentation task. 
Our findings indicate that including small objects with a large object scale yields optimal results. We surmise that objects spanning multiple scales offer more detailed insights and salient in-context samples lead to a richer diversity of information, which is beneficial for segmentation.

In our research on CA-ICL captioning, We explore the correlation between in-context input images and their corresponding descriptions. We drew inspiration from dense captioning and visual grounding, examining if incorporating object location information is beneficial for the model to capture semantic cues conveyed by in-context samples. 

As evidenced in Table~\ref{tab:ca_icl_caption}, introducing an image-type output leads to a notable decline in performance compared to the baseline. 
To tackle this issue, we explored the method of encoding bbox information in a textual format, as outlined in Section~\ref{sec:vl_prompt}. While the results were considerably better than the ``bbox\_image" approach, even outperformed the baseline in CIDEr metric. Figure~\ref{fig:ca_icl_cap_bbox} demonstrates that using prompts of the ``bbox\_text" type leads to more precise predicted captions that correspond with the intended region. This alignment significantly aids in the accurate and convenient verification of the model's performance during testing phases. This evidence supports the model's capability to effectively generate class-aware captions when supplied with appropriate examples.

\begin{figure}[t]
   \centering
   \includegraphics[width=0.7\linewidth,height=0.5\linewidth]{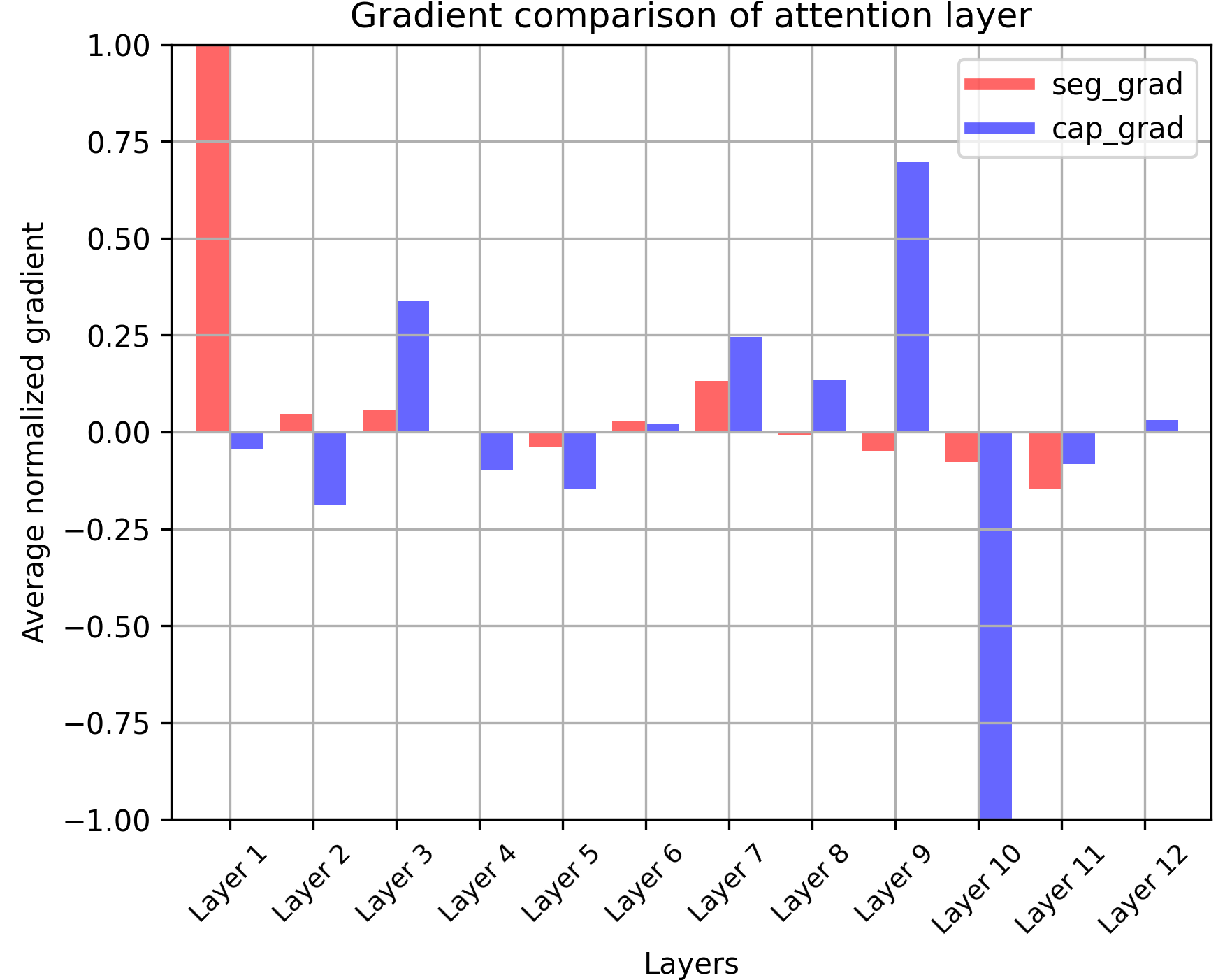}
   \vspace{-2mm}
   \caption{Gradient comparison of CA-ICL tasks. We utilize the normalized average gradient of each attention layer for comparison, while the symbol and the value represent the direction and magnitude of the gradient respectively.}
   \label{fig:avg_grad}
 \end{figure}

\vspace{0.5em}
\noindent \textbf{Model Variants Definition.}
We conducted experiments using various model configurations at a higher resolution of 256 to identify the optimal choice. The reference for these experiments is the single task performance, with the baseline established as task co-training using the standard GPT-2 small architecture, referred to as ``all tasks". We replace the FFN in part of transformer blocks with the MoE layer proposed in~\cite{Gshard} and the AG\_MoE introduced in~\cite{Uni-perceiver-moe} for analysis. The results presented in Table~\ref{tab:model_variants} reveal that the baseline setting results in significant unbalanced performance with a sharp segmentation performance decrease, while models with MoE configurations surpass the baseline in segmentation performance by 18.74 scores, yet there remains a notable shortfall of 10.8 scores in captioning performance. The adoption of the AG\_MoE structure further narrows this performance gap. Considering the image tokens dominate compared with text tokens and the divergent gradient directions of differing task complexities (as shown in Figure~\ref{fig:avg_grad}), the caption performance drops. Models with shared parameters might struggle to effectively manage the significant difference in token representations between the two tasks, highlighting the advantages of MoEs. In the following section, we will address the challenges associated with multi-task co-training.

\begin{table}[t]
\small
  \centering
  \resizebox{\linewidth}{!}{
  \begin{tabular}{lcc}
    \toprule
    \multirow{2.4}{*}{Model} & CA-ICL segmentation & CA-ICL captioning \\
    \cmidrule(lr){2-2} \cmidrule(lr){3-3}
    & MIoU \textcolor{red}{$\uparrow$} & CIDEr \textcolor{red}{$\uparrow$} \\
    \midrule
        \textcolor{gray}{single task}       & \textcolor{gray}{51.91}    & \textcolor{gray}{88.6}    \\
    \midrule
        all tasks                         & 21.74       & 77.3      \\
        {\scriptsize w/} MoE              & 40.48\resdiff{R}{+}{18.74}       & 66.5\resdiff{G}{-}{10.8}       \\
        {\scriptsize w/} AG\_MoE          & 42.02\resdiff{R}{+}{20.28}       & 67.9\resdiff{G}{-}{9.4}      \\ 
        {\scriptsize w/} MT               & 33.72\resdiff{R}{+}{11.98}       & 81.1\resdiff{R}{+}{3.8}     \\
        \rowcolor{blue!10} {\scriptsize w/} AG\_MoE and MT   & 49.91\resdiff{R}{+}{28.17}        & 78.3\resdiff{R}{+}{1.0}     \\
    \bottomrule
  \end{tabular}
  }
  \caption{Ablation of model variants and multi-task learning strategy. We present the MIoU and CIDEr metrics for CA-ICL segmentation and captioning tasks, respectively. In the brackets, we analyze gaps compared to the ``all tasks'' setting. We use \textcolor{G}{green} and \textcolor{R}{red} to indicate the performance decreases and increases.}
  \label{tab:model_variants}
  \vspace{-0.5em}
\end{table}

\begin{table*}[t]
  \small
  \setlength{\tabcolsep}{4pt}
  \centering
  \resizebox{0.85\textwidth}{!}{
  \begin{tabular}{lcccccccc}
  \toprule
   \multirow{2.4}{*}{Models} & \multirow{2.4}{*}{Resolution} & \multirow{2.4}{*}{\makecell[c]{\#Trainable \\ Params}} &  \multicolumn{2}{c}{CA-ICL Segmentation}  & \multicolumn{4}{c}{CA-ICL Captioning} \\
  \cmidrule(lr){4-5} \cmidrule(lr){6-9}
   & & & MIoU \textcolor{red}{$\uparrow$} & MAE \textcolor{red}{$\downarrow$} & B@4 \textcolor{red}{$\uparrow$} & METEOR \textcolor{red}{$\uparrow$} & CIDEr \textcolor{red}{$\uparrow$} & mAP \textcolor{red}{$\uparrow$} \\
    \midrule
    \multicolumn{9}{c}{specialist model} \\
    \midrule
    \textcolor{gray}{FPTrans~\cite{FPTrans}}       &\textcolor{gray}{480}       &\textcolor{gray}{139M}       &\textcolor{gray}{43.30}  &\textcolor{gray}{0.202}       &\textcolor{gray}{-}  &\textcolor{gray}{-}  &\textcolor{gray}{-}  &\textcolor{gray}{-} \\
    \textcolor{gray}{VAT~\cite{vat}}       &\textcolor{gray}{417}       &\textcolor{gray}{27M}       &\textcolor{gray}{46.07}  &\textcolor{gray}{0.087}       &\textcolor{gray}{-}  &\textcolor{gray}{-}  &\textcolor{gray}{-}  &\textcolor{gray}{-} \\
    \textcolor{gray}{DCAMA~\cite{DCAMA}}       &\textcolor{gray}{384}       &\textcolor{gray}{89M}       &\textcolor{gray}{53.06}  &\textcolor{gray}{\underline{0.059}}       &\textcolor{gray}{-}  &\textcolor{gray}{-}  &\textcolor{gray}{-}  &\textcolor{gray}{-} \\
    \textcolor{gray}{GRiT~\cite{GRiT}}       &\textcolor{gray}{1024}       &\textcolor{gray}{197M}       &\textcolor{gray}{-}  &\textcolor{gray}{-}       &\textcolor{gray}{5.2}  &\textcolor{gray}{9.0}  &\textcolor{gray}{58.6}  &\textcolor{gray}{\underline{15.9}}   \\
    \midrule
    \multicolumn{9}{c}{generalist model} \\
    \midrule
    SegGPT~\cite{Seggpt}       &448       &307M       &\underline{62.83}  &0.092       &-  &-  &-  &-   \\   
    SegGPT*       &256       &307M       &51.12  &0.116       &-  &-  &-  &-   \\
    OpenFlamingo~\cite{Openflamingo}       &224      &3B       &\textcolor{gray}{-}  &\textcolor{gray}{-}       &4.6  &11.4  &61.3  &-  \\
    \midrule 
    \textbf{Ours}       &256       &309M       & 58.04  & 0.110       &\textbf{5.3}  &\textbf{14.3}  &\textbf{86.9}  &10.9 \\ 
  \bottomrule
\end{tabular}
}
\caption{Comparison with state-of-the-art specialist and generalist models on class-aware in-context task. We report both the MIoU and MAE scores for comparison. * indicates that we test the SegGPT with images in 256 resolution. The previous state-of-the-art results are \underline{underlined}.}
\label{tab:main_results}
\end{table*}
 
\vspace{0.5em}
\noindent \textbf{Multi-task Co-training Strategy.}
In this section, we explore the impact of multi-task joint training. As demonstrated in Table~\ref{tab:model_variants}, employing the standard GPT-2 small architecture for co-training results in significant performance degradation, suggesting a considerable disparity in handling tasks involving different data modalities. The implementation of the AG\_MoE architecture results in a more balanced performance across tasks, yet there remains a notable performance gap compared to single-task scenarios. 

To further enhance the performance of the model with AG\_MoE, we adopt a multi-task learning paradigm to alleviate the task interference problems and, meanwhile, stabilize the training of MoEs. Drawing inspiration from Uni-Perceiver v2~\cite{Uni-perceiver2}, we utilize their unmixed batch sampling strategy and correlative optimizer. Here, the sampling weight $s_k$ of each dataset is configured to be proportional to the square root of the dataset's size. For the scaling factor $w_k$, we uniformly assign a value of 1 to all tasks. As evidenced in Table~\ref{tab:model_variants}, the integration of the AG\_MoE architecture with our multi-task learning strategy results in performance that exceeds the baseline for both tasks. This is particularly notable in the CA-ICL segmentation task, where an impressive gain of 28.17 points is observed. This indicates that the multi-task strategy effectively prevents potential task conflicts within a batch.

\subsection{Comparison with State-of-the-art Methods}
\label{sec:main_result}
We experimented with class-aware in-context tasks to compare with existing state-of-the-art specialist models as well as generalists. For the task definition, we adopt the best settings as discussed in ablations (Section~\ref{sec:ablation}). For the model and training strategy, we utilize AG\_MoE architecture with the multi-task learning strategy. 

For CA-ICL segmentation, we compare with generalist segmentation model SegGPT~\cite{Seggpt} and specialist few-shot segmentation models like FRTrans~\cite{FPTrans}, VAT~\cite{vat} and DCAMA~\cite{DCAMA}. As indicated in Table~\ref{tab:main_results}, our model trained at a resolution of 256 surpasses SegGPT that evaluated at the same resolution—an improvement of 6.92 in MIoU and 0.006 in the MAE score. However, still a gap between the 448 version of SegGPT with more training data and higher resolution input. The performance is also notably comparable to the state-of-the-art specialist DCAMA, which operates at a higher resolution of 384 as well.

In the domain of CA-ICL captioning, the generalist baseline for evaluation is Openflamingo~\cite{Openflamingo}, a large vision-language model that excels in demonstrating strong in-context captioning ability. The CA-ICL captioning task most analogous to it is that of dense captioning, as both tasks necessitate the prediction of not only the caption but also the corresponding bbox. Therefore, we compare with the sota dense captioning model GRiT~\cite{GRiT}. We utilize the images in our test set to evaluate GRiT. Then allocate the generated predictions to our ground-truth regions annotations, utilizing the IoU metric of their respective bboxes as the basis for the assignment. As shown in Table~\ref{tab:main_results}, our method achieves state-of-the-art performance in traditional image captioning metrics. In comparison to Openflamingo, which has a parameter tenfold greater, we also achieve a 0.7-point increase in BLEU4 and a significant 25.6-point improvement in CIDEr. However, the result still has a gap in the mAP score compared with GRiT. We believe this is because they utilize a foreground object extractor. 
\section{Conclusion}
In this work, we present a unified framework for in-context visual understanding. By leveraging multimodal quantization and unified embedding, our model is capable of jointly learning multimodal data in the general token embedding space. 
By synergistically integrating autoregressive transformer with the MoEs framework, we achieve stable multi-task co-training while simultaneously benefiting from the balanced contributions of each task. 
Overall, our research showcases the potential of in-context learning across various modalities as well as tasks.

\clearpage
\clearpage
\setcounter{page}{1}
\maketitlesupplementary

\begin{table}[ht]
  \small
  \setlength{\tabcolsep}{2.5pt}
  \centering
  \resizebox{\linewidth}{!}{
    \begin{tabular}{lcc}
        \toprule
        Hyperparameter & GPT-2 & Ours \\
        \midrule
        Architecture & transformer decoder & transformer decoder \\
        Vocabulary size & 50257 & 51290 / 52295 \\
        Max positions & 1024 & 2060 \\
        Hidden size & 768 & 768 \\
        Hidden layers & 12 & 12 \\
        Attention heads & 12 & 12 \\
        Number of MoEs & - & 6 \\
        Number of experts & - & 8 \\
        Layer norm epsilon & 1e-05 & 1e-12 \\
        Attention probs dropout prob & 0.1 & 0.1 \\
        Hidden dropout prob & 0.1 & 0.1 \\
        \bottomrule
    \end{tabular}
}
\caption{Hyperparameters for our GPT-2 baseline and proposed model. Note for the vocabulary size, we experiment with two settings for whether to add the special category and bbox tokens.}
\label{tab:hyperparameters}
\end{table}

\begin{table}[ht]
  \small
  \setlength{\tabcolsep}{3.5pt}
  \centering
  \resizebox{0.75\linewidth}{!}{
  \begin{tabular}{lcccccc}
  \toprule
   &  \multicolumn{3}{c }{B@4 \textcolor{red}{$\uparrow$}} & \multicolumn{3}{c}{CIDEr \textcolor{red}{$\uparrow$}} \\
  \cmidrule(lr){2-4} \cmidrule(lr){5-7}
   Examples  & 1 & 2 & 3 & 1 & 2 & 3 \\
    \midrule 
    $\mathcal{L}_{out}$         &\textbf{5.4}  &1.8  &1.6       &\textbf{95.9}  &52.5  &51.2    \\
    \midrule
    {\scriptsize w/} $\mathcal{L}_{in}$      &4.5  &1.9  &1.9       &85.0  &\textbf{54.7}  &54.5    \\
    {\scriptsize w/} $0.5\mathcal{L}_{in}$   &5.3  &\textbf{2.0}  &\textbf{2.0}       &86.6  &54.6  &\textbf{55.6}    \\
  \bottomrule 
  \end{tabular}
}
\caption{Loss analysis on class-aware in-context captioning task.}
\label{tab:loss_analysis_cap}
\end{table}

\begin{table}[h]
  \small
  \setlength{\tabcolsep}{2.5pt}
  \centering
  \resizebox{\linewidth}{!}{
  \begin{tabular}{cccccc}
  \toprule
  \multirow{2}{*}{Examples} & \multicolumn{2}{c}{CA-ICL segmentation} & \multicolumn{3}{c}{CA-ICL captioning} \\
  \cmidrule(lr){2-3} \cmidrule(lr){4-6}
  &  MIoU\textbf{ \textcolor{red}{$\uparrow$}}  & MAE\textbf{ \textcolor{red}{$\downarrow$}} &  B@4\textbf{ \textcolor{red}{$\uparrow$}} & CIDEr\textbf{ \textcolor{red}{$\uparrow$}} & mAP\textbf{ \textcolor{red}{$\uparrow$}} \\
    \midrule
    0     &45.70  &0.094     &2.8  &65.7  &0.2 \\
    \midrule
    \multicolumn{6}{c}{w/o category information for CA-ICL captioning} \\
    \midrule
    1     &56.17  &0.167     &6.6  &95.6  &1.5 \\
    2     &59.21  &0.132     &1.7  &45.5  &1.7 \\
    3     &60.85  &0.128     &0.8  &32.6  &1.6 \\
    \midrule
    \multicolumn{6}{c}{w category information for CA-ICL captioning} \\
    \midrule
    1     &58.04  &0.110     &5.3  &86.9  &10.9 \\
    2     &61.65  &0.101     &2.3  &60.9  &0.8 \\
    3     &62.33  &0.098     &2.3  &62.2  &1.5 \\
  \bottomrule
\end{tabular}
}
\caption{Analysis on in-context samples and category information. We report the metrics utilized in our main experiments for the two CA-ICL tasks.}
\label{tab:icl_effectiveness}
\end{table}


\section{Model Architecture and Configuration}

We provide more details on the model architecture compared with GPT-2 small as shown in Table~\ref{tab:hyperparameters}, where each dense decoder layer is the same and the even-numbered layer is replaced with the sparse decoder layer as discussed in Section 3.3 of our main submission. Another notable difference is the vocabulary size, We employ two different settings: one that includes special category and bbox tokens, and another without them. Compared with GPT-2, we adopt a smaller layer norm epsilon of 1e-12 to ensure stable training.

\begin{figure}[t]
   \centering
   \includegraphics[width=\linewidth]{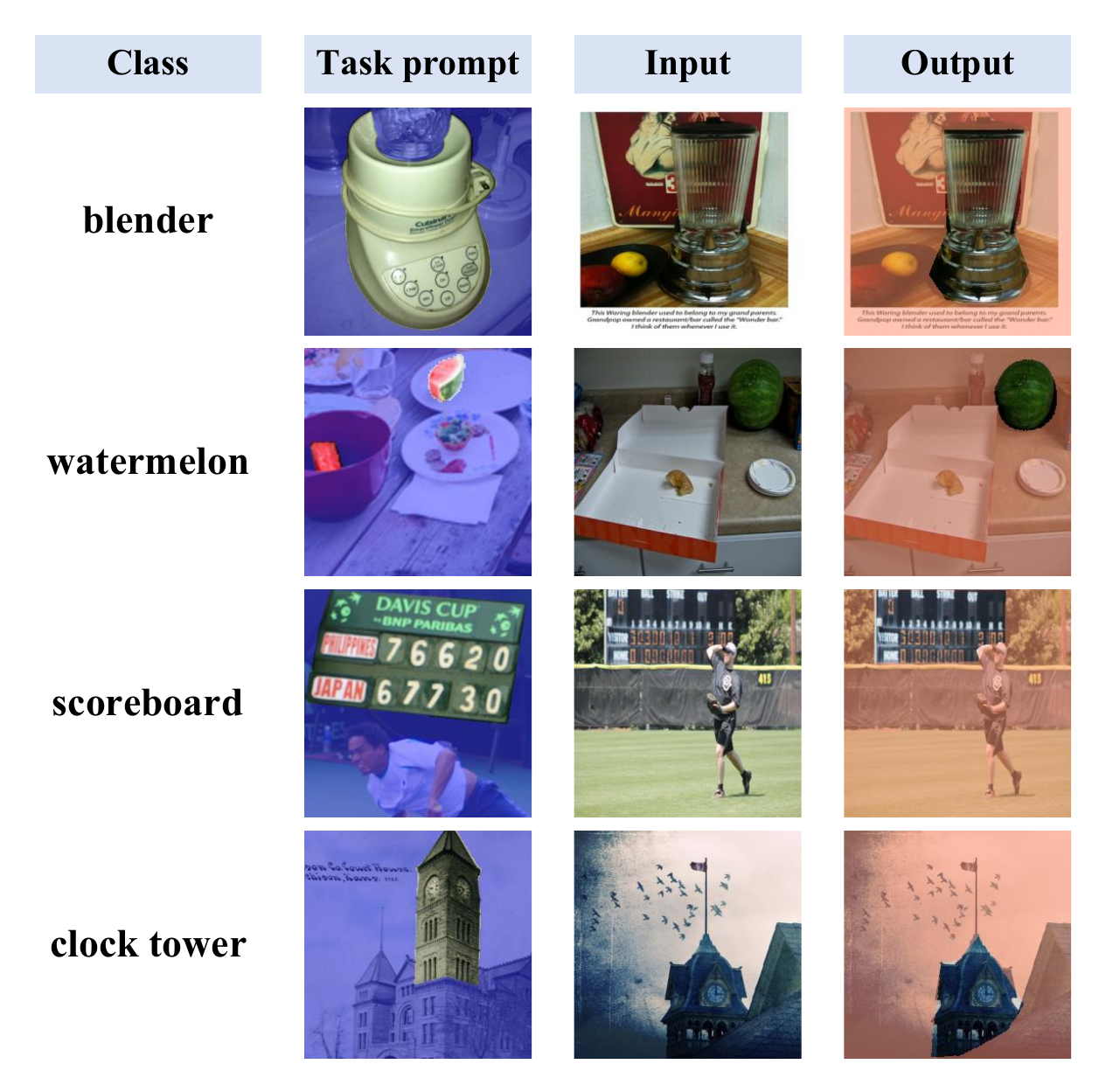}
   \caption{Out-of-domain test in CA-ICL segmentation. We employ the non-overlapping classes of the LVIS dataset to create a per-class mask pool, following the same approach used with the MS-COCO dataset.}
   \label{fig:seg_vis_ood_1}
 \end{figure}

\begin{figure}[t]
   \centering
   \includegraphics[width=\linewidth]{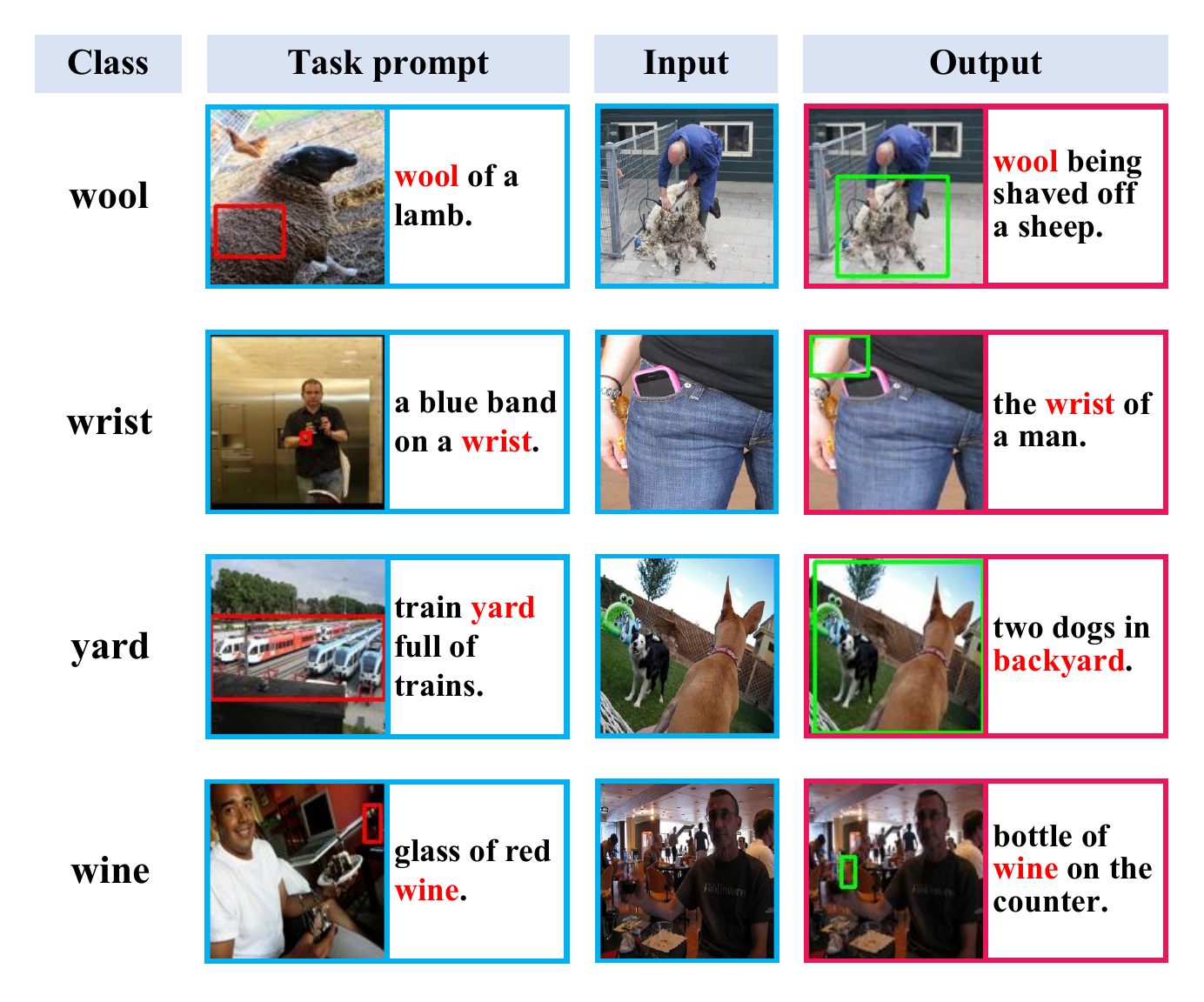}
   \caption{Out-of-domain test in CA-ICL captioning. We employ the non-overlapping classes of the Visual Genome dataset to create a per-class pool, following the same approach as discussed in Section 4.1 of our main submission.}
   \label{fig:cap_vis_ood_1}
 \end{figure}

\section{Additional Quantative Analysis}

\noindent \textbf{Loss and padding analysis for CA-ICL captioning.}
In this section, we delve into the effects of loss functions and padding strategies on captioning performance. We employ the baseline loss $\mathcal{L}_{out}$, delineated in Section 3.3 of our main submission, aligning the length of caption tokens with image tokens via padding. We examine different CE loss weights for the input image tokens in each context (denoted as $\mathcal{L}_{in}$). This setting is based on the intuition that image captioning task may benefit from an increased focus on visual content because of the unbalanced sequence length between the image and text tokens. As indicated in Table~\ref{tab:loss_analysis_cap}, the experiment results reveal that using only $\mathcal{L}_{out}$ surpasses other configurations with $\mathcal{L}_{in}$ in a one-shot setting. In contrast, a composite loss of $0.5\mathcal{L}_{in} + \mathcal{L}_{out}$ achieves superior results in two- and three-shot scenarios. Consequently, we adopt the $0.5\mathcal{L}{in} + \mathcal{L}{out}$ loss for individual captioning tasks, while utilizing a consistent $\mathcal{L}_{out}$ during co-training sessions.

\noindent \textbf{In-context effectiveness analysis.}
We study the impact of increasing the number of in-context pair examples and whether to add category information in CA-ICL captioning in the task prompt on the outcomes. We trained the model using three in-context samples and inference with 1 to 3 samples. Additionally, we trained one model with explicit class information input instead of in-context samples for comparison. As presented in Table~\ref{tab:icl_effectiveness}, with only class information, the model performs pool on both tasks for the CA-ICL segmentation task, which indicates the effectiveness of in-context samples as they are given more information than the simple class label. The inclusion of more examples consistently improves the segmentation performance. However, for the CA-ICL captioning task, the performance does not exhibit a steady increase, even more serious if category information is not provided. The possible reason is that using more in-text samples for the segmentation task can provide more segmentation clues coming from different views, and appearances of different image samples for the same category of target object. But for the caption task, one caption is already enough to denote the target object while multiple description styles from different samples will introduce more style ambiguity. From the perspective of performance, we report the best results of the model with class information for the captioning task. The problem of captioning is left to further study.

\noindent \textbf{Time cost of different models.}
We calculate the fps metric for the 1 in-context example setting to analyze the time cost. The inference speed of our model using 0, 1, 2, 3 in-context examples is 2.8, 2.4, 1.9, 1.4 fps, respectively. While for SegGPT and OpenFlamingo, the fps is 7.7 and 0.3 img/s. Our model is capable of using in interactive applications.

\begin{figure*}[t]
   \centering
   \includegraphics[width=\linewidth,height=0.95\textheight]{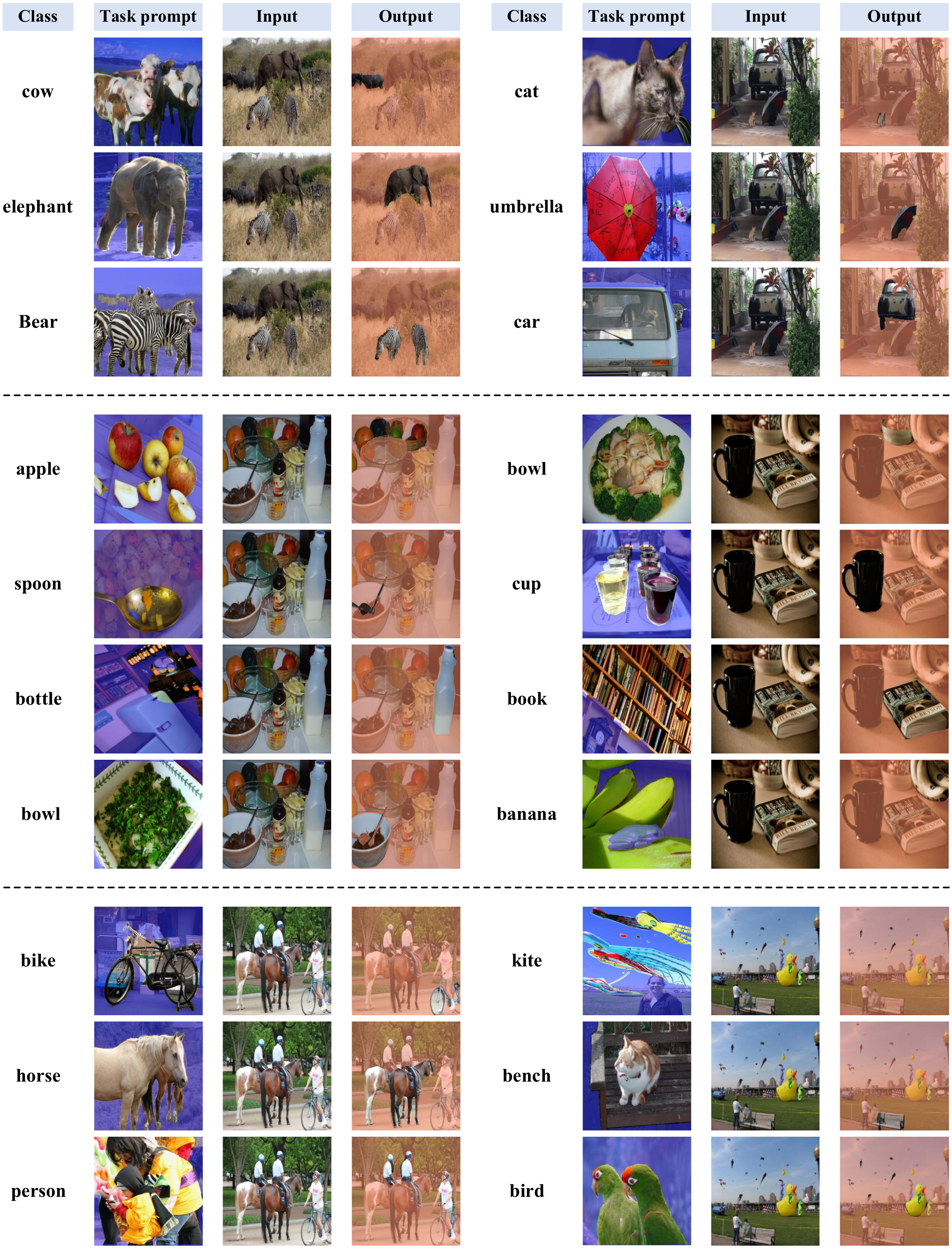}
   \caption{Results of CA-ICL segmentation. Our model demonstrates robustness across in-context prompts, effectively handling objects from diverse classes and accommodating variations in size and quantity. For better visualization, we overlay the mask onto the corresponding image. In this setup, the \textcolor{blue}{blue} area indicates the mask for in-context prompts, while the \textcolor{red}{red} area represents the output mask.}
   \label{fig:seg_vis_1}
 \end{figure*}

\begin{figure*}[t]
   \centering
   \includegraphics[width=\linewidth]{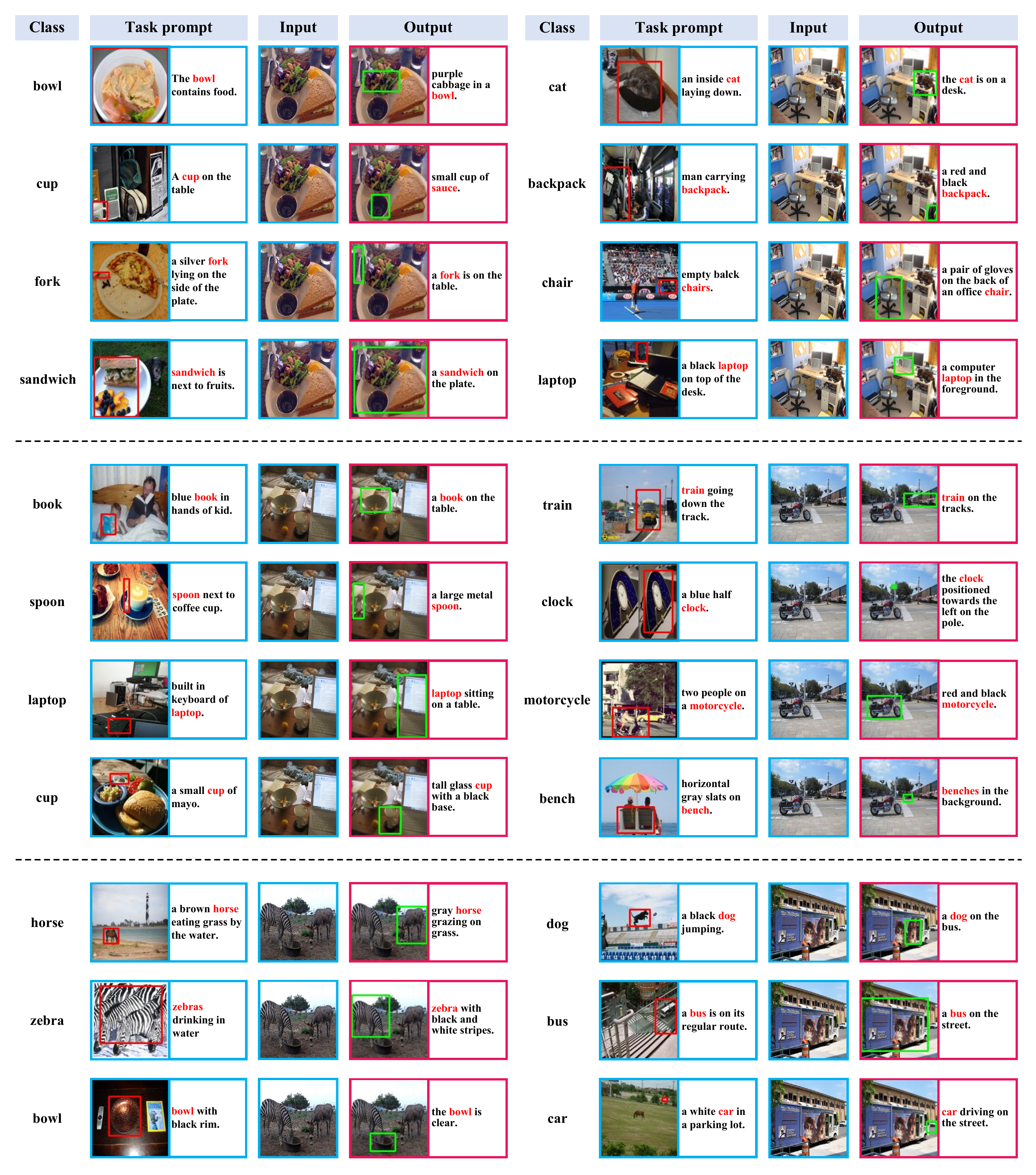}
   \caption{Qualitative results of CA-ICL captioning. Our model shows great semantic reasoning, accurately interpreting clues within in-context samples. It generates relatively precise bounding boxes and descriptions that correspond well with the desired objects in the images. To enhance visual clarity, we illustrate bounding boxes in in-context samples using red squares \textcolor{red}{$\square$}, and the predicted bounding boxes are marked in green \textcolor{green}{$\square$}. Additionally, we emphasize category information in the captions by using \textcolor{red}{red} text.}
   \label{fig:cap_vis_1}
 \end{figure*}

\begin{figure*}[t]
   \centering
   \includegraphics[width=0.85\linewidth]{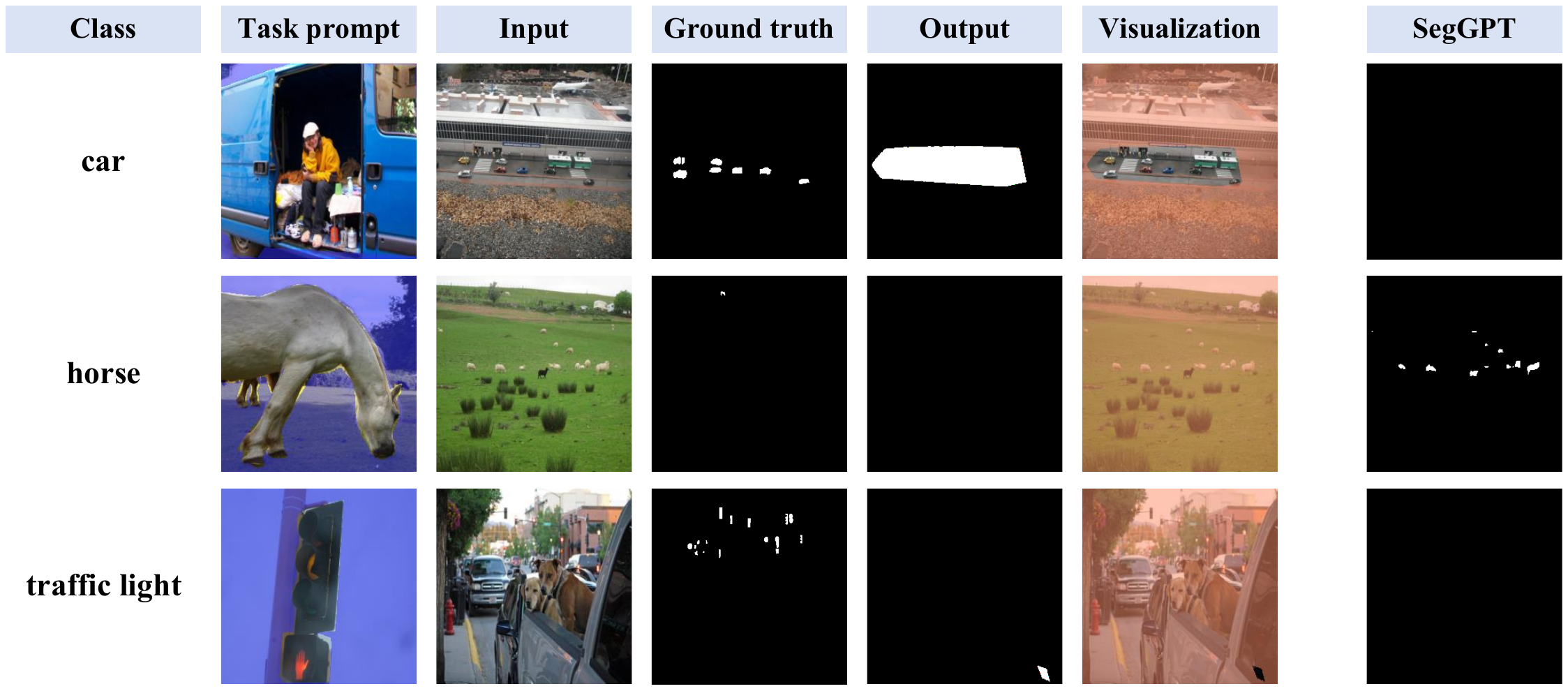}
   \caption{Typical failure case for CA-ICL segmentation. The model faces challenges in processing input images with numerous small instances and also performs worse with categories that are infrequently represented in the training data. Zoom in for a better view.}
   \label{fig:fail_seg_vis_1}
 \end{figure*}

\begin{figure*}[t]
   \centering
   \includegraphics[width=\linewidth]{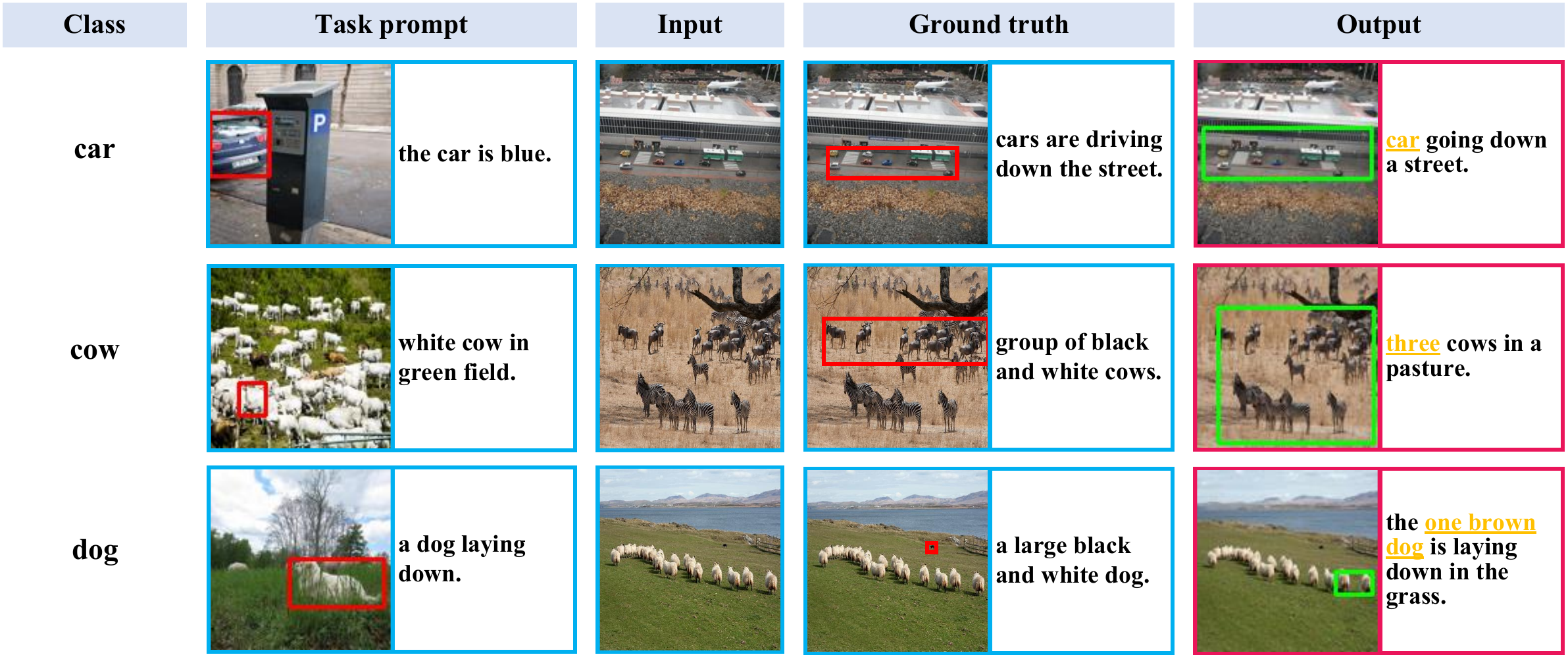}
   \caption{Typical failure case for CA-ICL captioning. When facing multiple instances or small instances, the model may predict inaccurate region location or wrong caption as highlighted in \textcolor[RGB]{255,192,0}{yellow}. Zoom in for a better view.}
   \label{fig:fail_cap_vis_1}
 \end{figure*}

\section{Additional Qualitative Results}
\noindent \textbf{In-context reasoning.}
To illustrate the in-context reasoning ability of our model, we provide qualitative results on the two CA-ICL tasks. As illustrated in Figure~\ref{fig:seg_vis_1}, the model shows excellent semantic understanding for both in-door and out-door scenarios. Given suitable input prompts, the model demonstrates exceptional reasoning capabilities in segmenting instances that belong to the same category as the in-context samples. 
Figure~\ref{fig:cap_vis_1} showcases the model's ability to generate accurate captions with locations that precisely identify the region of the desired category, demonstrating its strong reasoning capabilities as well.

\noindent \textbf{Out-of-domain tests.}
To evaluate the efficacy of semantic clues and the model's capabilities, we conducted out-of-domain tests on two distinct tasks. As illustrated in Figure~\ref{fig:seg_vis_ood_1}, the model demonstrates proficiency in utilizing cues from in-context examples featuring categories not encountered during training, thereby achieving dependable segmentation results. Additionally, for the captioning task, we utilized a per-category pool derived from the Visual Genome dataset, specifically selecting category data that do not coincide with the training categories. The results shown in Figure~\ref{fig:cap_vis_ood_1} further revealed the model's ability to generalize effectively to unfamiliar categories.

\section{Limitation and future work}
Because of the long-tailed class and object scale distribution of the training dataset, the model does not perform well with multiple small objects or uncommon classes like traffic light.
Some typical failure cases are presented in Figure~\ref{fig:fail_seg_vis_1} and Figure~\ref{fig:fail_cap_vis_1}. We think a more balanced data distribution may be beneficial for the situation. For example, utilizing Copy-Paste strategies~\cite{copy_paste,x_paste} to expand the per-category instance pool. For improving captioning, a potential solution involves resampling the data or other data balancing strategies. Another limitation is that the model only supports one class per forward. Currently, the proposed model can support multiple categories by multiple times inference. The color mapping strategy utilized in SegGPT might be helpful. 

The proposed method can accommodate a more diverse range of in-context learning tasks beyond the scope of class-aware tasks. As we discussed in Section~\ref{sec:uni-repre}, the multi-modal input will be quantized and mapped into the unified representation space. Therefore, all modal inputs quantized by modality-specific quantizers can be modeled using our framework, regardless of the task. Next, we plan to extend M$^2$oEGPT to accommodate even more modalities (\eg, web page, 3D vision, heat map, tables) and tasks (\eg, image generation and editing, inpainting, and grounding), also support high-resolution image and longer output, broadening the system's applicability such that it becomes more general. 




{
    \small
    \bibliographystyle{ieeenat_fullname}
    \bibliography{main}
}


\end{document}